\pdfoutput=1

\documentclass[11pt]{article}

\usepackage[final]{acl}
\usepackage{times}
\usepackage{latexsym}

\usepackage[T1]{fontenc}

\usepackage[utf8]{inputenc}

\usepackage{microtype}

\usepackage{inconsolata}

\usepackage{graphicx}
\usepackage{amsmath}
\usepackage{amsfonts}
\usepackage{enumitem}
\usepackage{multirow}
	
\newcommand\MNAME{AMR-RE}
\newcommand\NONREFTNAME{\textit{Unsupervised}}
\newcommand\REFTNAME{\textit{Supervised}}
\newcommand\PENAME{SAP}
\newcommand\GENAME{SAP+CTX}

\newcommand\AMRSIMNAME{SS-GNN}
\title{\textcolor{blue}{\MNAME}: \textcolor{blue}{A}bstract \textcolor{blue}{M}eaning \textcolor{blue}{R}epresentations for Retrieval-Based In-Context Learning in \textcolor{blue}{R}elation \textcolor{blue}{E}xtraction}

\author{
Peitao Han\textsuperscript{1}, Lis Kanashiro Pereira\textsuperscript{2},  Fei Cheng\textsuperscript{3},
Wan Jou She\textsuperscript{4}, Eiji Aramaki\textsuperscript{1}
 \\ 
  \textsuperscript{1} Nara Institute of Science and Technology, Japan  \\
  \textsuperscript{2} National Institute of Information and Communications Technology (NICT), Japan\\
  \textsuperscript{3} Kyoto University, Japan  \\
  \textsuperscript{4} Kyoto Institute of Technology, Japan \\
  { han.peitao.hr3@is.naist.jp, 
 liskanashiro@nict.go.jp, feicheng@i.kyoto-u.ac.jp}\\ {  wjs2004@kit.ac.jp, aramaki@is.naist.jp}
}

\begin{document}
\maketitle

\begin{abstract}

Existing in-context learning (ICL) methods for relation extraction (RE) often prioritize language similarity over structural similarity, which may result in overlooking entity relationships. We propose an AMR-enhanced retrieval-based ICL method for RE to address this issue. Our model retrieves in-context examples based on semantic structure similarity between task inputs and training samples. We conducted experiments in the \REFTNAME~ setting on four standard English RE datasets. The results show that our method achieves state-of-the-art performance on three datasets and competitive results on the fourth. Furthermore, our method outperforms baselines by a large margin across all datasets in the more demanding \NONREFTNAME~setting.

\end{abstract}
\section{Introduction}



Large language models (LLMs) exhibit strong in-context learning (ICL) abilities across various NLP tasks simply by being given a few examples of the task. However, the quality of few-shot demonstrations can substantially impact the performance of ICL, and tasks requiring high precision, such as relation extraction, remain challenging. 


Relation extraction (RE) is a task to identify a predefined semantic relation between entity pairs mentioned in the context. Relations between entity pairs are often implicitly expressed, which can lead to suboptimal ICL performance. Existing ICL methods for RE often overlook the semantic associations between entity pairs, relying primarily on entity mentions or overall sentence semantics for representation (\citealp{han2023information}; \citealp{wan-etal-2023-gpt}; \citealp{ijcai2024p704}; \citealp{ma-etal-2023-chain-thought}; \citealp{sun2023pushing}).

Abstract Meaning Representation (AMR) \cite{banarescu2013abstract} provides a detailed semantic graph structure that represents semantics through nodes and edges, where nodes correspond to semantic elements such as events, entities and arguments, and edges indicate the relationships between them. 
AMR graphs offer precise descriptions of entities by incorporating their arguments and semantic roles, making them well suited for the RE task (\citealp{hu-etal-2023-semantic}; \citealp{zhang-ji-2021-abstract}; \citealp{gururaja-etal-2023-linguistic}).

\begin{figure*}[t]
  \includegraphics[width=\linewidth]{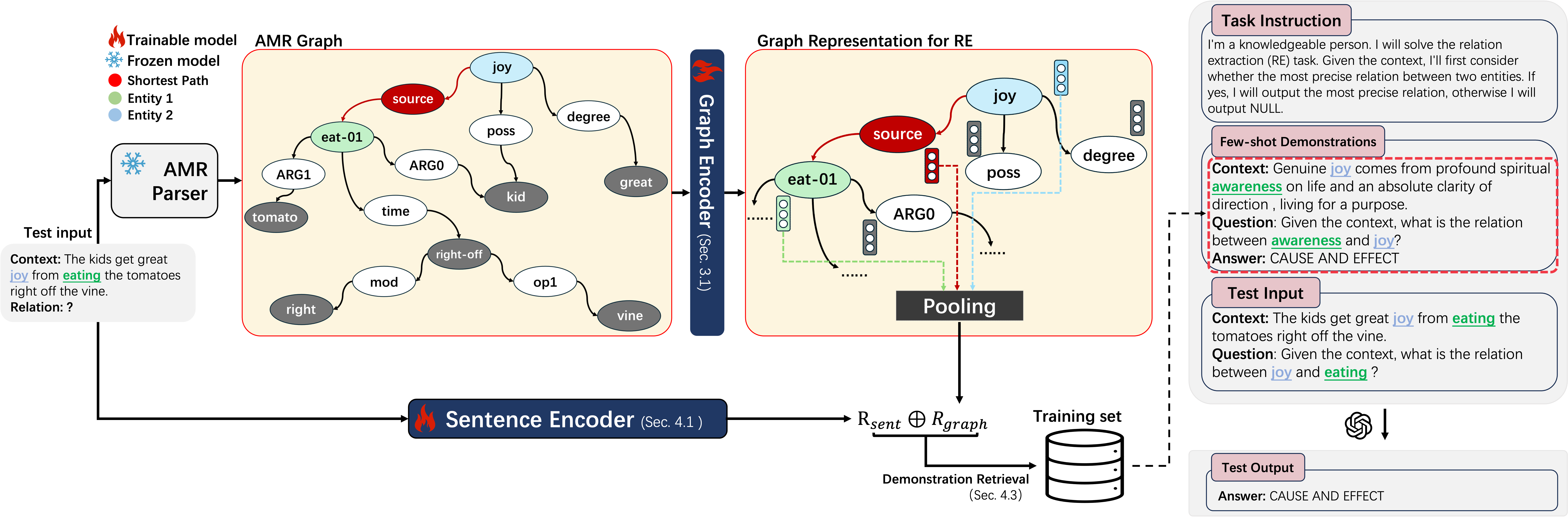} 
  \caption {\textbf{An overview of our proposed method in the \REFTNAME~ Setting (Section \ref{sec:model}, Section \ref{sec:sup_retri}).} Given a test input, we first adopt our AMR-enhanced demonstration retrieval method to select proper demonstrations from the training set. Subsequently, all retrieved demonstrations are included in the prompt construction.}
  \label{fig:main}
\end{figure*}

As shown in Figure \ref{fig:main}, the input sentence, "... get great \underline{joy} from \underline{eating} ...", is parsed into a semantic graph, where the node "source" connects to two entity nodes ("joy" and "eat-01"). This structure explicitly represents the Cause-Effect relation between these two arguments, illustrating how semantic graphs can capture underlying relational meanings beyond surface text.


To bridge the contextual gap caused by missing semantic structure,
we propose \textit{\textbf{\MNAME}}, an AMR-enhanced retrieval-based ICL method that leverages AMR graphs to select in-context examples based on semantic structure similarity. Evaluations on four English RE datasets show that our method surpasses state-of-the-art methods on three datasets with the \REFTNAME~ AMR-based retriever (Section \ref{sec:sup_retri}). To comprehensively assess our approach, we further evaluate \MNAME~in the more challenging \NONREFTNAME~setting. Our simple yet effective architecture (Section \ref{sec:unsup_retri}) consistently achieves higher F1 scores compared to sentence embedding-based ICL baselines.

\section{Preliminaries}

\subsection{Task Definition}
Given a set of pre-defined relation classes $\mathbb{R}$, relation extraction aims to predict the relation $y \in \mathbb{R}$ between the given pair of subject and object entities $(e_{sub}, e_{obj})$ within the input context $\mathbf{C}$, or if there is no pre-defined relation between them, predict $y = \text{NULL}$. We formalize RE as a language generation task, and introduce the prompt construction in the next section.

\subsection{Prompt Construction\label{prompt}}
We construct a prompt for each test example. Each prompt consists of three components:

    
    \noindent\textbf{Instructions}: We provide a precise description of the RE task and a set of pre-defined relation classes  $\mathbb{R}$. The model is required to output the relation corresponding to these predefined classes; if the relation does not belong to any of these classes, the model will output NULL.
    
    \noindent\textbf{ICL Demonstrations}: Given one test example, we search k-Nearest Neighbor ($k$NN) demonstrations via two different frameworks: \textit{\REFTNAME} (Section \ref{sec:sup_retri}) and \textit{\NONREFTNAME} (Section \ref{sec:unsup_retri}). All demonstrations are included in the prompt.

    \noindent\textbf{Test Input}: We provide the test input in the same format as the ICL Demonstrations, and the LLM is expected to output the relation.

\section{The \MNAME~Model}
\label{sec:model}

This section gives an overview of our \textit{\textbf{\MNAME}}~method (Figure \ref{fig:main}).
Given an input text, \MNAME~first generates its AMR graph using an off-the-shelf AMR parser. A self-supervised graph model then encodes this graph to obtain the graph embeddings. These embeddings are then used to retrieve $k$NN examples from the training set for ICL~\cite{wan-etal-2023-gpt}.

Our method leverages the shortest path between two entities for retrieving RE demonstrations, as it aligns with the core objective (supplying semantic structure) of the RE task.


\subsection{AMR Graph Encoding}
\label{sec:graph_enc}
\noindent\textbf{AMR Graph Construction:} To generate the AMR graph from the input text,  we adopt an off-the-shelf AMR parser\footnote{\url{https://github.com/IBM/transition-amr-parser}}. We parse the input sentence into an AMR graph $G=\{V,E,R\}$, where $V$, $E$, $R$ are the sets of nodes, edges, and relation types, respectively. In $G$, the edge labeled $(u,r,v) \in E$, where $u, v \in V$ and $r \in R$, means that there is an edge labeled $r$ from node $u$ to node $v$.


\noindent\textbf{Self-supervised Graph Encoder:} After constructing the AMR graph from the input text, we use a graph encoder to produce the graph embeddings. 
\citet{shou-lin-2023-evaluate} employ a \underline{\textbf{s}}elf-\underline{\textbf{s}}upervised approach to train an AMR  \underline{\textbf{g}}raph-based  \underline{\textbf{n}}eural  \underline{\textbf{n}}etwork; this model assesses the AMR similarity through the encoded representations, hereafter referred to as the \textit{\AMRSIMNAME} model. We adapt \textit{\AMRSIMNAME} for the RE task by optimizing it on our proposed graph RE representations. Notably, this training framework only depends on the corpus without annotated relation labels. This method explicitly optimizes representations by assessing the similarity between two AMR graphs via a contrastive loss. Training details are added in Appendix \ref{sec:appendix_amrsim}.
Given an AMR graph $G=[(u_1,r_1,v_1),\cdots,(u_n,r_n,v_n)]$, $G$ is linearized by a depth-first traversal algorithm $G=[u_1,r_1,v_1,\cdots,u_n, r_n,v_n;\mathcal{A}]$, where $\mathcal{A}$ is the adjacency matrix. $G$ will be fed to \AMRSIMNAME~to obtain the node representations $H_{node} = \{h_{node}^{u_1},h_{node}^{r_1},\cdots,h_{node}^{v_n}\}$ where $h_{node}^{a}$ denotes as the node representation of node $a$.
\begin{equation}
H_{node} = \text{\AMRSIMNAME}([u_1,r_1,v_1,\cdots,v_n]; \mathcal{A})
\end{equation}

\subsection{Graph Representation for RE}
The \textit{\AMRSIMNAME} model originally employs mean pooling of all nodes in the AMR graph as the graph representation, which is also used for self-supervised training. While this approach has demonstrated significant advancements in overall AMR similarity assessment, it is not optimized for identifying relationships between two specific entities. To address this limitation, we construct graph RE representations specifically designed for RE, focusing on capturing the structural and semantic information of entities and their relationships.

Inspired by previous works, the shortest path between two entities in the semantic structure \cite{hu-etal-2023-semantic} or the syntactic structure \cite{cheng-miyao-2017-classifying} often contains crucial information needed to determine relations. Based on these insights, we focus on leveraging the shortest AMR path (SAP) as the most informative subgraph for retrieving the relevant RE demonstrations.

To investigate the optimal way of representing a relation with AMR graph representations for the RE task, we establish fine-grained setups for the graph RE representation $R_{graph}$. Typically, the shortest path between the entity pair $(e_{obj},e_{sub})$ can be denoted as $V_{path}=\{e_{obj},p_1,p_2,\cdots,p_n,e_{sub}\}$ where $V_{path} \in V$, and $p_i$ represents intermediate nodes on the shortest AMR path. We investigated two different pooling strategies and two path modeling strategies.


\noindent\textbf{Pooling Strategy:} To analyze the impact of the pooling strategy on $R_{graph}$, we adopt two pooling methods:

(1) \textit{Mean Pooling}: We use the average of all node representations from the shortest path for retrieval, formally $R_{graph}=\frac{1}{|V_{path}|} \sum_{v_i \in V_{path}} h_{node}^{v_i}$.

(2) \textit{Concatenation}: The node representations of the entity pair, $h_{node}^{e_{obj}}$ and $h_{node}^{e_{sub}}$, are concatenated with the mean pooling of the nodes along the shortest AMR path to form the final graph representation, formally $R_{graph}=h_{node}^{e_{obj}} \oplus h_{node}^{e_{sub}} \oplus h_P$, where $h_P=\frac{1}{n} \sum_{i=1}^{n} h_{node}^{p_i}$.

\noindent\textbf{Path Modeling:} We use two distinct methods to explore how to effectively leverage information from the shortest path:

(1) \textit{\PENAME}: This approach strictly isolates all the information from the components not in the shortest path between entity nodes, and only the shortest AMR path is fed to \textit{\AMRSIMNAME}, which encodes the node representations along the path. The final graph RE representation $R_{graph}$ is constructed by pooling the node representations within the path.

(2) \textit{\GENAME}: We use the whole AMR graph as the input for \textit{\AMRSIMNAME}. In this setup, the node representations benefit from bidirectional attention and the GNN adapter, allowing them to integrate contextual information from neighbor nodes. The pooling of the node representations within the shortest AMR path is then formed as the graph RE representation.

By combining the pooling and path modeling strategies, we obtained four distinct configurations, with detailed results provided in Table \ref{table:allresults}.

\section{AMR-Based Demonstration Retrieval}
In this section, we introduce two settings for incorporating AMR graph information to retrieve ICL demonstrations. First, we present the \REFTNAME~setting, where \MNAME~benefits from both graph and sentence RE representations (Section \ref{sec:sup_retri}). To further evaluate the effectiveness of our method, we assess \MNAME~under the more challenging \NONREFTNAME~setting (Section \ref{sec:unsup_retri}). \MNAME~retrieves in-context examples by $k$NN retrieval from the training set using the relation representation $R_{rel}$ (Section \ref{sec:retrieval}).

\subsection{\textit{\REFTNAME}~Setting}
\label{sec:sup_retri}

In the \REFTNAME~setting, we integrate both sentence-level and structural information to achieve optimal performance and explore the potential interactions between these two types of representations. 

\noindent\textbf{Sentence RE Representations:} We use PURE \cite{zhong-chen-2021-frustratingly}, an entity marker-based RE model. For example, given the input sentence “And \underline{we} will see \underline{you} then”, the subject entity "we" and object entity "you", the sentence becomes: “[CLS] And [SUB\_ORG] \underline{we} [/SUB\_ORG] will see [OBJ\_PER] \underline{you} [/OBJ\_PER] then [SEP]”. The final hidden representations of the BERT encoder are denoted as $H_{sent} = \{h_{sent}^{1},\cdots,h_{sent}^{m}\}$ where $h_{sent}^{i}$ denotes the $i$-th hidden representation. Let $s_{obj}$ and $s_{sub}$ be the indices of the beginning of the entity markers 
[SUB\_ORG] and [OBJ\_PER]. 
We define the sentence representation as $R_{sent} = h_{sent}^{s_{obj}} \oplus h_{sent}^{s_{sub}}$, where $\oplus$ denotes the concatenation of representations along the first dimension.

\noindent\textbf{Graph RE Representations:} We obtain graph RE representations $R_{graph}$ from the \AMRSIMNAME~as we introduced in Section \ref{sec:graph_enc}.

We use the concatenation of AMR graph embeddings $R_{graph}$ from \AMRSIMNAME~ and sentence embeddings $R_{sent}$ from PURE, formally $R_{rel}=R_{graph}\oplus R_{sent}$. \AMRSIMNAME~and PURE are fine-tuned on RE datasets by predicting the relation probability from $R_{rel}$ through a feedforward network. Notably, \AMRSIMNAME~is first self-supervised trained, then subsequently fine-tuned on RE task.

\subsection{\textit{\NONREFTNAME}~Setting}
\label{sec:unsup_retri}

We further evaluate our approach in the more challenging \NONREFTNAME~ setting for comprehensively analyzing the effectiveness of AMR graph. 
In this setting, \MNAME~ retrieves examples using only graph RE representations $R_{graph}$, which means $R_{rel}=R_{graph}$. Note that \AMRSIMNAME~ is only self-supervised on the corpus without annotated relation labels in \NONREFTNAME~setting. We compare our model with Sentence RE Representations-based baselines.

\subsection{Demonstration Retrieval}
\label{sec:retrieval}
The relation representation $R_{rel}$ is used to perform $k$NN retrieval, where the top-$k$ most similar demonstrations are selected and included in the prompt. To efficiently implement $k$NN demonstration retrieval, we adopt FAISS \citep{johnson2019billion} library for efficient search.



\section{Experiments} \label{sec:exp}

\begin{table*}
  \small
  \centering
  \begin{tabular}{l|c|c|c|c|c|c}
    \hline
     \textbf{Method} &\textbf{Retriever}  &\textbf{SemEval ($\Delta$\%)} & \textbf{TB-DENSE ($\Delta$\%)} & \textbf{SciERC ($\Delta$\%)}  & \textbf{ACE05 ($\Delta$\%)}&\textbf{Avg}  \\
    \hline
    \multicolumn{7}{c}{\REFTNAME~ Setting} \\
    \hline
    PURE &- & 90.77  & 66.70  & 67.08  &  \textbf{68.62}  & 73.57 \\
    GPT-RE\_FT &\textit{PURE} & 91.46   & 67.58   & 67.32   &  68.59   & 73.74\\ 
    \MNAME~(Ours)  &\scriptsize{\textit{\AMRSIMNAME+PURE} }      &  \textbf{\underline{91.97} ($\uparrow$ 0.6)}  & \textbf{71.54 ($\uparrow$ 5.9)} & \textbf{$\mathbf{68.10^{\ast}}$ ($\uparrow$ 1.1)} & $67.94^{\ast}$ ($\downarrow$ 0.9)    & \textbf{74.89}  \\
    

    \hline
        \multicolumn{7}{c}{\NONREFTNAME~ Setting} \\
    \hline
    GPT-Random  &-   & 67.83 &  22.03  &16.48  &9.73 & 29.02                     \\ 
    GPT-Sent   &SimCSE      & 77.64   &28.73   &21.60  &10.04 & 34.50                 \\
    GPT-RE\_Entity+  &SimCSE        &  80.25  & 31.19   & 26.15    &13.10    & 37.67                   \\
    
    \MNAME~(Ours) &\textit{\AMRSIMNAME}    &  \textbf{\underline{84.68} ($\uparrow$ 5.5)}  & \textbf{38.17 ($\uparrow$ 22.4)} &\textbf{$\mathbf{27.89^{\ast}}$ ($\uparrow$ 6.7)} & \textbf{$\mathbf{15.04^{\ast}}$ ($\uparrow$ 14.8)}   & \textbf{41.45}  \\
    \hline
  \end{tabular}
  \caption{\label{table:mainresults}
    \textbf{Main results.} We set the number of demonstrations to $k=10$. For \MNAME, we only report the best results from the four distinct configurations obtained by combining the pooling and path modeling strategies, explained in Section \ref{sec:graph_enc} (see Table \ref{table:allresults} for detailed results). Underlined results refer to the \textit{\PENAME} graph RE representation, otherwise, \textit{\GENAME} is applied. The $\Delta$\% indicates the corresponding differences in percentage when compared to GPT-RE\_FT and GPT-RE\_Entity+ in \REFTNAME~and \NONREFTNAME~settings respectively. The Avg column shows the average score for all datasets. The highest results are in \textbf{bold}. $\ast$ denotes that this result is implemented by concatenation pooling, otherwise, mean pooling is used.
  }
\end{table*}

\noindent \textbf{Backbone LLM:} We use OpenAI’s GPT-4 as the LLM model in \MNAME~and in all baselines, and we set the number of demonstrations to $k=10$ in the main results. For a fair comparison, all results are reproduced by ourselves. Baselines such as \citet{wan-etal-2023-gpt} originally used GPT-3.5 (text-davinci-003), however, this model is not available through the OpenAI API anymore. In addition, GPT-4 has been shown to outperform its previous versions in several NLP tasks and was the SOTA backbone for ICL at the time. Our method can be easily applied to other backbones as well, however, models such as Llama currently cannot match GPT-4’s performance in ICL \citep{chatterjee-etal-2024-language}.

\noindent \textbf{Evaluation Datasets:}
We evaluate our model on four English RE datasets. Two general domain RE datasets: SemEval 2010 Task 8 \citep{hendrickx-etal-2010-semeval} and ACE05\footnote{\url{https://catalog.ldc.upenn.edu/LDC2006T06}}, one temporal RE dataset: TimeBank-Dense \citep{cassidy-etal-2014-annotation}, and one scientific domain dataset: SciERC \citep{luan-etal-2018-multi}. Due to the high cost of the OpenAI API, following \citet{wan-etal-2023-gpt}, we sample a subset of ACE05 dataset (due to its large size) for our experiments. 
Details of each dataset are provided in Appendix \ref{sec:appendix_datasets}. We adopt Micro-F1 as evaluation metrics.
The hyperparameter settings are provided in the Appendix \ref{sec:appendix_hyperparam}.
\section{Main Results}

\subsection{Results in the \REFTNAME~Setting}

\noindent \textbf{Baselines in \REFTNAME~Setting:} To analyze the effectiveness of the AMR graph, we select two baseline methods for comparison with \MNAME.

(1) \REFTNAME~RE Baseline w/o ICL: We implement \textit{PURE} \citep{zhong-chen-2021-frustratingly} as a directly comparable baseline to show the impact of ICL. 

(2) Baseline with \REFTNAME~Retrievers: We implement \textit{GPT-RE\_FT} \citep{wan-etal-2023-gpt} as the baseline with a \REFTNAME~retriever. GPT-RE\_FT employs representations encoded by PURE \citep{zhong-chen-2021-frustratingly}. 

\noindent \textbf{Results:} Table \ref{table:mainresults} shows our results. Overall, \MNAME~ outperforms the baselines in the \REFTNAME~ setting. This indicates that the more explicit representation of AMR graphs enhances the quality of the retrieved demonstrations. 
In the \REFTNAME~setting, \MNAME~achieves SOTA performance on the SemEval, SciERC and TB-Dense datasets while delivering competitive results on the ACE05 dataset.
The results indicate that the fine-tuned structure representation benefits from both structural and semantic information. However, ACE05 contains a large proportion of the samples annotated as NULL relation, which introduces significant noise. This can mislead the model during both retriever training and ICL inference, resulting in decreased performance compared to the fully-supervised baseline, PURE.



\subsection{Results in the \NONREFTNAME~Setting}
\noindent \textbf{Baselines in \NONREFTNAME~Setting:} We select three baselines that are comparable to \MNAME~ in \NONREFTNAME~setting. The details of each baseline are introduced below:

(1) \textit{GPT-Random:} we randomly select few-shot ICL demonstrations with additional constraints to ensure a more uniform label distribution; 

(2) \textit{GPT-Sent:} we follow \citet{bioiclre} to retrieve $k$NN demonstrations with SimCSE~\cite{gao-etal-2021-simcse}, which is a widely used sentence embedding model;

(3) \textit{GPT-RE\_Entity+:} we adopt the entity-prompted sentence embedding proposed by \citet{wan-etal-2023-gpt} that incorporates both the entity pair and contextual information for retrieval.

\noindent \textbf{Results:} Table \ref{table:mainresults} shows our results in the \NONREFTNAME~setting. \MNAME~consistently outperforms the baselines on all four datasets. These findings underscore the efficacy of AMR-enhanced graph RE representations in effectively capturing relational information. In particular, by focusing on the shortest AMR path, \MNAME~highlights core entities and the semantic relations between them, thereby reducing noise and providing clearer relational cues compared to conventional sentence-embedding-based approaches.


\begin{table}
\small
  \centering
  \begin{tabular}{l|c|c}
    \hline
    \textbf{Method} &\textbf{SemEval ($\Delta$\%)} & \textbf{SciERC ($\Delta$\%)}\\
    \hline
    \multicolumn{3}{c}{Supervised Setting} \\ \hline
   \MNAME & \textbf{91.97} & \textbf{68.10}
    \\
    \quad\textit{w/o self-sup} & 90.82 ($\downarrow$ 1.3) & 67.04 ($\downarrow$ 1.6) \\
    \quad\textit{w/o $R_{sent}$} & 89.71 ($\downarrow$ 2.5) & 67.19 ($\downarrow$ 1.3) \\
    \quad\textit{w/o $R_{graph}$} & 91.46 ($\downarrow$ 0.6) & 67.32 ($\downarrow$ 1.2)\\
    \hline
    \multicolumn{3}{c}{Unsupervised Setting} \\ \hline
    \MNAME & \textbf{84.68 } & \textbf{27.56} \\
    \quad\textit{w/o self-sup } & 81.67 ($\downarrow$ 3.6) & 26.01 ($\downarrow$ 5.6)\\  \hline
  \end{tabular}
  \caption{\label{table:assst}
    \textbf{Ablation study.} For the full model, we show the best configuration results from Table \ref{table:mainresults}.  \textit{w/o self-sup} indicates that the retriever is not self-supervised on the target dataset. The $\Delta$\% is the percentage of corresponding difference.
  }
\end{table}
\section{Ablation Study}

\begin{figure}[t]
  \includegraphics[width=\columnwidth]{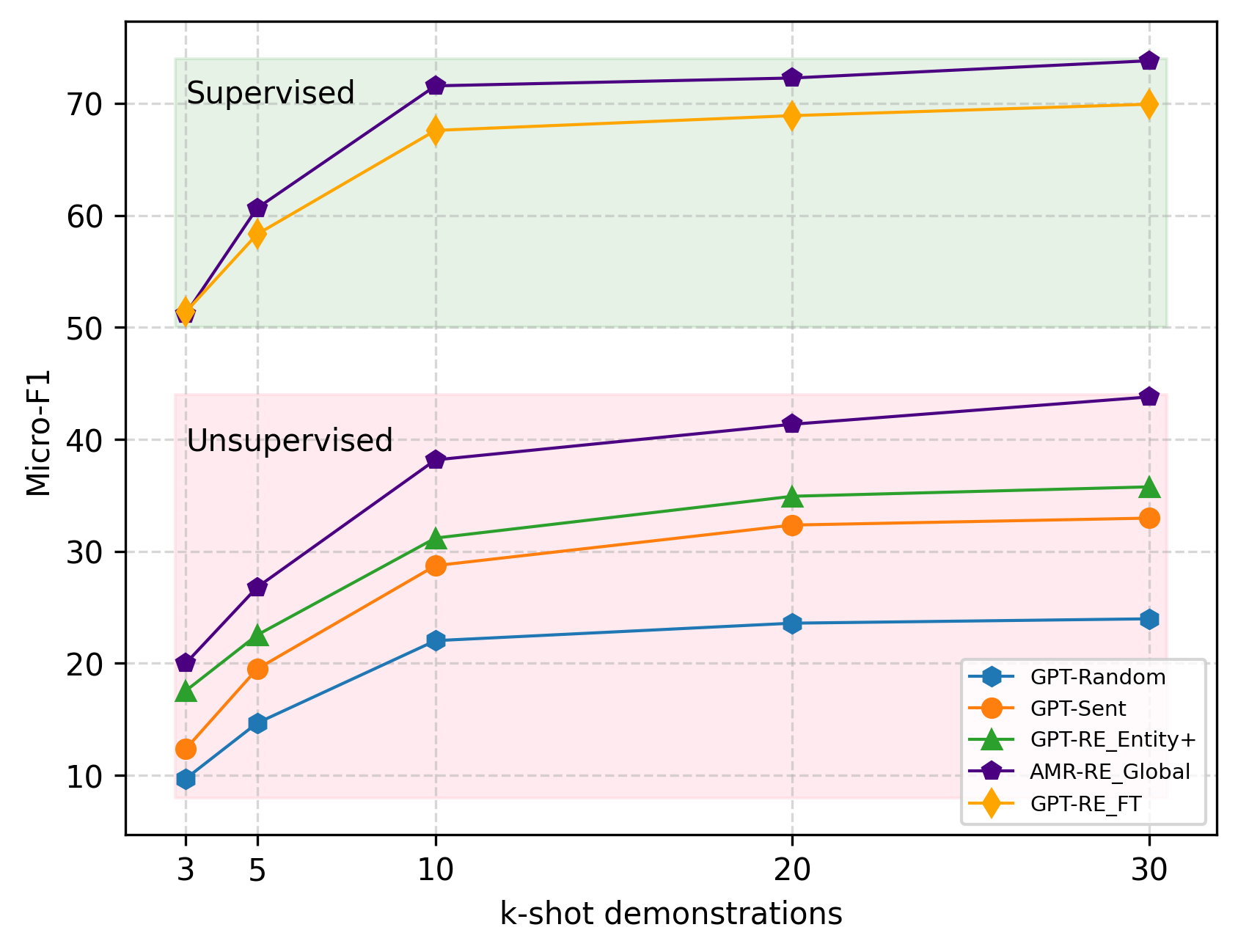 }
  \caption{Performance for the different number of few-shot examples on TB-Dense.}
  \label{fig:kshot}
\end{figure}

Table \ref{table:assst} illustrates the impact of self-supervision on the graph encoder and the roles of sentence and graph RE representations in the relation representations. The results show that self-supervision enhances performance, with graph ($R_{graph}$) and sentence ($R_{sent}$) representations both being crucial in the \REFTNAME~setting.
We also investigated the impact of the number of demonstrations on performance. Figure \ref{fig:kshot} shows that \MNAME~consistently outperforms the baselines across all $k$-shots, demonstrating the effectiveness of incorporating AMR graphs for retrieval.

\section{Case Study}

\begin{figure}[t]
  \includegraphics[width=\columnwidth]{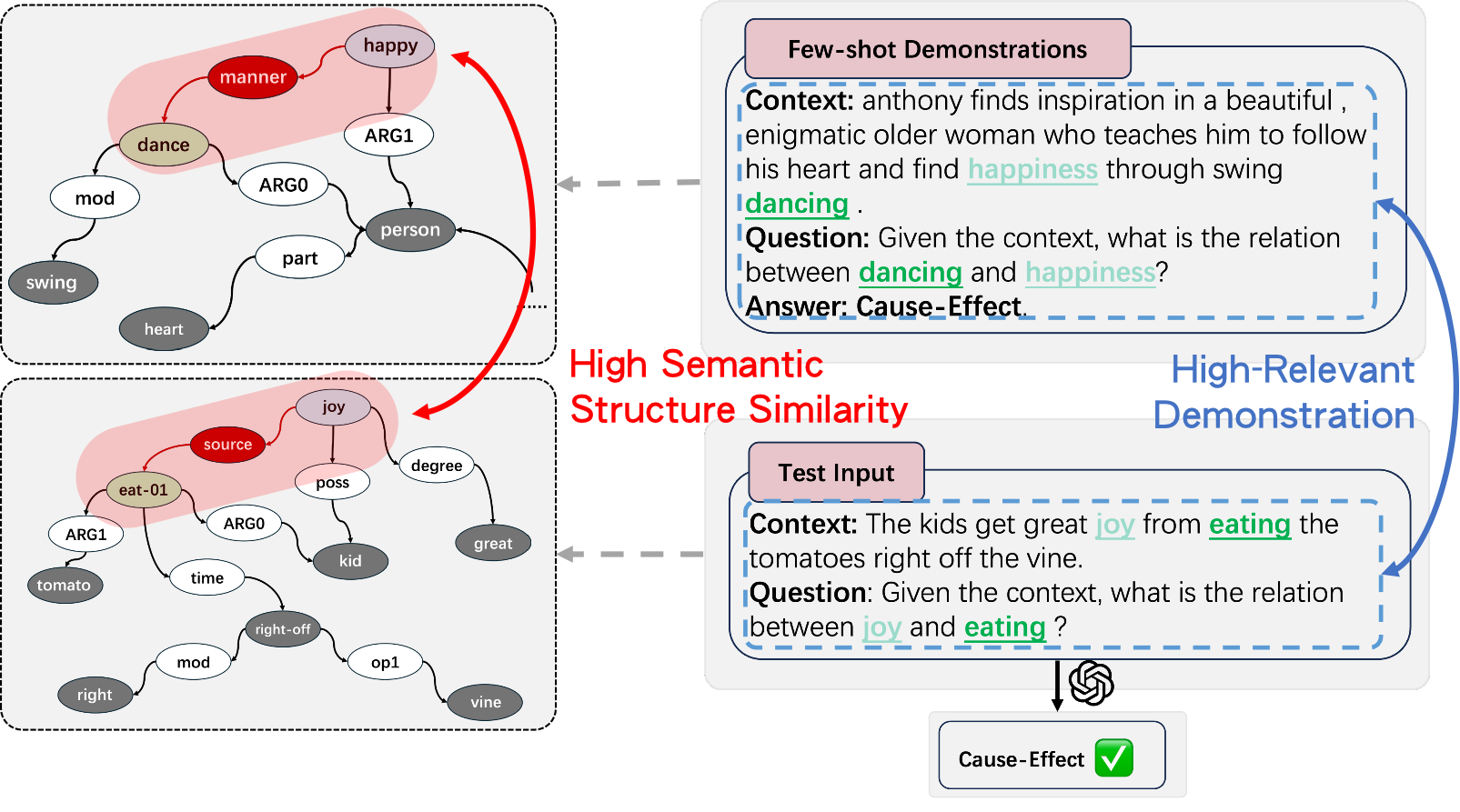 }
  \caption{A case study of semantic structure similarity. The demonstration with similar semantic structure enables the LLM to correctly generate the gold label, "Cause-Effect".}
  \label{fig:cases1}
\end{figure}

\begin{figure}[t]
  \includegraphics[width=\columnwidth]{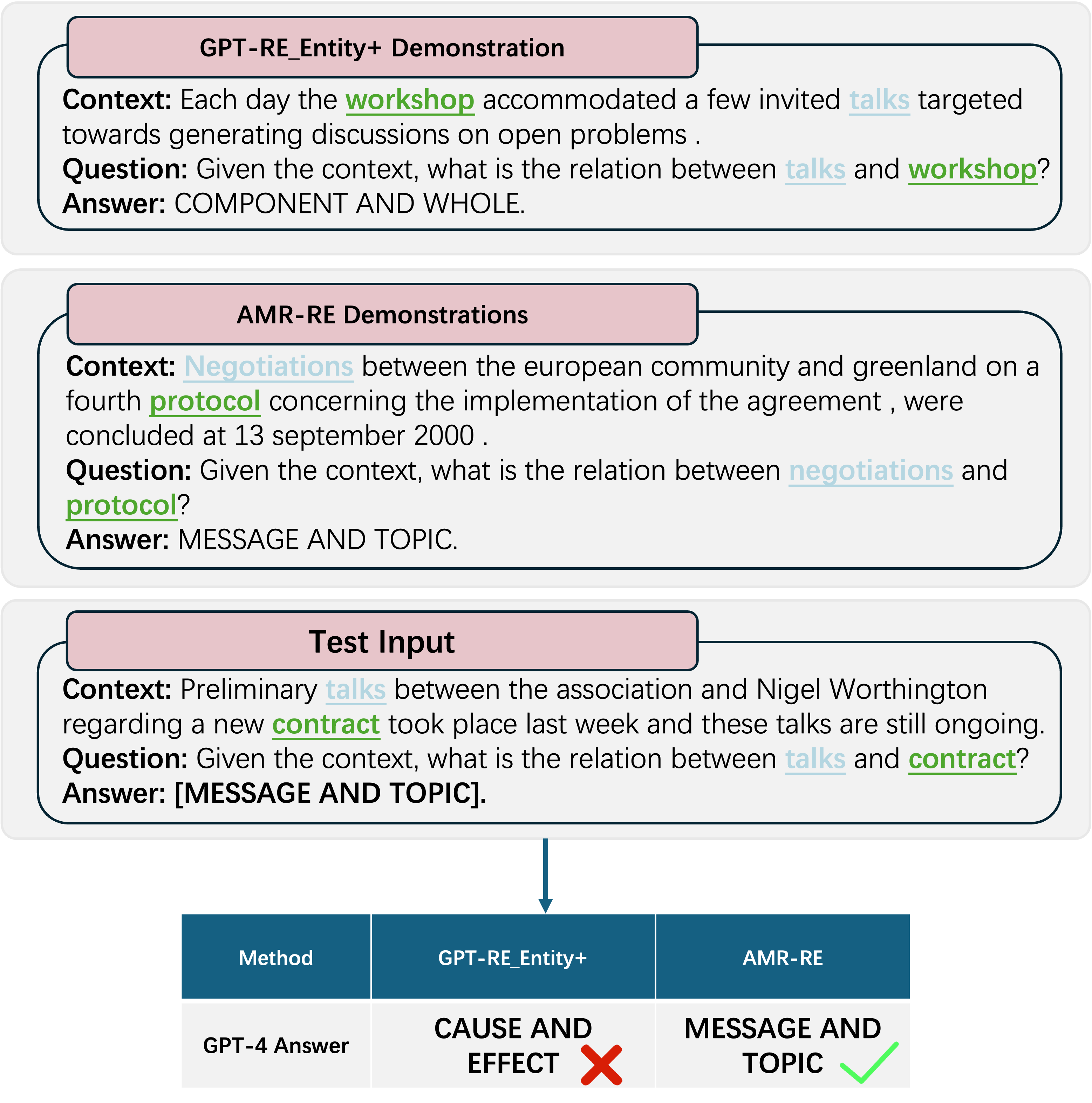 }
  \caption{A case study of \textbf{\MNAME}~retrieved demonstration quality. MESSAGE AND TOPIC is the gold label.}
  \label{fig:cases2}
\end{figure}

To demonstrate how semantic structure similarity enables the retrieval of highly relevant demonstrations and surpasses sentence-based baselines on RE ICL, we present two representative case studies in the \NONREFTNAME~Setting.
Figure \ref{fig:cases1} illustrates that our proposed AMR enhanced retrieval method effectively captures both the similarity of event structure and the semantics of the entities. This shows that demonstrations with high semantic structure similarity serve as more suitable and informative RE demonstrations for ICL.
Figure \ref{fig:cases2} highlights the effectiveness of \textbf{\MNAME}. Our proposed method successfully retrieves few-shot RE demonstrations with semantically equivalent entities (e.g., "protocol"–"contract", "negotiations"–"talks"), while also capturing implicit relational connections. It demonstrates \MNAME's ability to align both explicit and implicit semantic information for improved relation extraction. In contrast, the sentence-based retrieval method fails to model such information.

\section{Conclusions}


We proposed \textit{\textbf{\MNAME}}, an AMR-enhanced retrieval-based ICL method that uses AMR graphs to select demonstrations based on semantic structure similarity. Evaluations on four English RE datasets show that \MNAME~outperforms the baselines. This underscores the effectiveness of combining graph learning with LLMs for relation extraction. Our experiments further demonstrate that AMR graph information can lead to more accurate and robust relation extraction, even in \NONREFTNAME~settings.

\section{Limitations}

We focused our work on:
1) demonstrating the effectiveness of graph similarity in retrieval-based ICL on the RE task. However, our work can be generalized beyond RE, as AMR is a universal semantic analysis tool applicable to other tasks, and ICL is also not restricted to RE; 2) evaluating our method on English RE datasets, mainly because AMR parsers only offer promising performance in English \cite{cai-etal-2021-multilingual-amr}. There are other semantic tools, such as multilingual dependency parser \cite{ustun-etal-2020-udapter}, for constructing graphs that extend beyond English. 
\bibliography{anthology,custom}

\appendix
\section{Self-supervised Training for AMR Graph Encoding}
\label{sec:appendix_amrsim}

\AMRSIMNAME~ \cite{shou-lin-2023-evaluate} adopts a self-supervised approach, Contrastive Tension (CT) to optimize the representation of an AMR graph. The main assumption is that AMR graphs with adjacent distributions have similar meanings. In our work, we adapt this approach to our novel AMR graph representation.


Two independent transformer-based encoders that also incorporate graph neural networks are identically initalized. The training objective is to maximize the dot product between positive pairs $(G_{p},G_{p}^{+})$ while minimizing the dot product between negative pairs $(G_{p},G_{p}^{-})$. For each randomly selected AMR graph $G_{p}$, we use $G_{p}^{+}=G_{p}$ to create a positive pair. Then, we construct negative instances by pairing $G_p$ with $K$ randomly sampled different graphs. The $K+1$ instances are included in the same batch. The training contrastive loss $\mathcal{L}$ is binary cross-entropy between similarity scores and labels.
\begin{equation}
\mathcal{L} =
\begin{cases} 
-\log \sigma(h_{graph}\cdot h_{graph}^{+})\\
-\log \sigma(1-h_{graph}\cdot h_{graph}^{-})
\end{cases}
\end{equation}

where $\sigma$ refers to the Logistic function; $h_{graph}$ is the graph representation. 
The model is then updated to compute the similarity between the two graphs.

\begin{table}
\small
  \centering
  \begin{tabular}{l|c}
    \hline
   \textbf{Hyperparameter} & \textbf{Value}\\
    \hline
   Engine Name & GPT-4-0314  \\
   Temperature & 0 \\
   Top\_P & 1 \\
   Frequency\_penalty & 0 \\
   Presence\_penalty & 0 \\
   Best\_of & 1 \\ \hline
  \end{tabular}
  \caption{\label{table:gpt_setting}
    \textsf{GPT-4} hyperparameters.
  }
\end{table}

\section{Evaluation Datasets}
\label{sec:appendix_datasets}

\begin{table*}
\small
  \centering
  \begin{tabular}{l|c|c|c|c}
    \hline
   \textbf{Dataset} &\textbf{\# Relation} & \textbf{\# Train} & \textbf{\# Dev} & \textbf{\# Test (\# Subset)} \\
    \hline
   SemEval & 9 & 6,507 & 1,493 & 2,717 (2,717)  \\
   TB-Dense & 6 & 7,553 & 898 &  2,299 (2,299) \\
   SciERC & 7 & 16,872 & 2,033 & 4,088 (4,088) \\
   ACE05 & 6 & 121,368 & 27,597 & 24,420 (2,442) \\ \hline
  \end{tabular}
  \caption{\label{table:data_stat}
   Statistics of the evaluation datasets. \# Subset denotes the number of instances sampled from the original test set, due to the high cost of the OpenAI API.
  }
\end{table*}

In this section, we describe the evaluation datasets used in our experiments. Table \ref{table:data_stat} shows the statistics
for each dataset.

\noindent\textbf{SemEval 2010 Task 8} \citep{hendrickx-etal-2010-semeval}: This data set focuses on the semantic relations between pairs of nominals. It was annotated from general domain resources. The task is to classify the semantic relations into one of nine directed relation types: Cause-Effect, Instrument-Agency, Product-Producer, Content-Container, Entity-Origin, Entity-Destination, Component-Whole, Member-Collection, Message-Topic, and Other (to indicate that there is
no relation between the pair of nominals). An example of a sentence with an event pair that holds the Cause-Effect relation is shown below:

\begin{quote}
The \textbf{(e1:discomfort)} from the \textbf{(e2:injury)} was now precluding him from his occupation which involved prolonged procedures in the standing position.
\end{quote}

\noindent\textbf{ACE05}: This dataset contains entities, relations, and events annotated from resources from domains including newswire, broadcast news, broadcast conversation, weblog, discussion forums, and conversational telephone speech. It requires identifying semantic relations into the following six types: artifact, general-affiliation, organization-affiliation, part-whole, person-social, physical. The following example contains an entity pair with the part-whole relation:

\begin{quote}
Witnesses say they heard blasts around a presidential complex in the \textbf{(e1:center)} of the \textbf{(e2:city)}.
\end{quote}

\noindent\textbf{TB-Dense} \cite{cassidy-etal-2014-annotation}: TB-Dense is a public benchmark for
temporal relation extraction (TRE). It was annotated from TimeBank \cite{pustejovsky2003timebank} and TempEval \cite{uzzaman-etal-2013-semeval}. We use a preprocessed version from \cite{wang-etal-2022-maven} for experiments. TB-Dense annotates temporal relations for event pairs within adjacent sentences. To handle this, we separately parse the two sentences into AMR graphs and then connect the two graphs through a shared root node following \citep{cheng-miyao-2017-classifying}. Given a passage and two event points, the task is to classify the relations between events into one of six types: BEFORE, AFTER, SIMULTANEOUS, VAGUE, IS\_INCLUDED, and INCLUDES. An example with two events, e1 and e2 (in bold) that hold the SIMULTANEOUS relation is shown below:

\begin{quote}
Nobody \textbf{(e1:hurried)} her up.	No one \textbf{(e2:held)} her back.
\end{quote}

\noindent\textbf{SciERC} \citep{luan-etal-2018-multi}: This dataset includes annotations
for scientific entities and their relations annotated from 500 scientific abstracts taken from Artificial Intelligence conferences and workshops proceedings. The relation types are: \textit{used-for}, \textit{feature-of}, \textit{hyponym-of}, \textit{part-of}, \textit{compare}, \textit{conjunction} and \textit{corefence}. Following example contains the \textit{feature-of} relation between two entities: 

\begin{quote}
They improve the reconstruction results and enforce their consistency with a \textbf{(e1:priori knowledge)} about \textbf{(e2:object shape)}.
\end{quote}

\section{Hyperparameters}
\label{sec:appendix_hyperparam}

\begin{table*}
  \small
  \centering
  \begin{tabular}{l|c|c|c|c|c|c|c}
    \hline
    \textbf{Setting}  &\textbf{Path} &\textbf{Pooling}  &\textbf{SemEval} & \textbf{TB-DENSE} & \textbf{SciERC}  & \textbf{ACE05}&\textbf{Avg}  \\
    \hline
    \multirow{4}{*}{\REFTNAME}
    &\multirow{2}{*}{SAP+CTX} &\textit{Mean} &  90.84  & \textbf{71.54} & 67.92 & 67.37   & 74.22  \\ 
     & &\textit{Concatenation}       &  90.03  & 70.56 & \textbf{68.10}& \textbf{67.94}    & 74.36 \\
      
    &\multirow{2}{*}{SAP} &\textit{Mean}        &  \textbf{91.97} & 68.23 & 67.81 & 66.80   & 73.70  \\
     & &\textit{Concatenation}       &  91.70  & 67.89 & 68.04 & 67.21    &   73.71 \\
    \hline
    \multirow{4}{*}{\NONREFTNAME}
    &\multirow{2}{*}{SAP+CTX} &\textit{Mean} &  81.40  & \textbf{38.17} & 27.64 & 14.82   & 40.51  \\ 
    &  &\textit{Concatenation}       &  79.48  & 37.78 & \textbf{27.89}& \textbf{15.04}    & 40.05 \\
    &\multirow{2}{*}{SAP} &\textit{Mean}       &  \textbf{84.68} & 35.64 & 27.56 & 14.65    & 40.63  \\
    &  &\textit{Concatenation}      &  83.51  & 33.75 & 27.61 & 14.69    &   39.89 \\
    \hline
  \end{tabular}
  \caption{\label{table:allresults}
    \textbf{\MNAME~results with all configurations.} The results in \textbf{bold} are reported in the main results.
  }
\end{table*}

\noindent\textbf{GPT-4}:
We used GPT-4 by the OpenAI API \footnote{\url{https://platform.openai.com/docs/api-reference/introduction}} during the experiments. The hyperparameters used can be found in Table \ref{table:gpt_setting}, we report the result of the single run for all experiments.

\noindent\textbf{Unsupervised Sentence Embedding Model}: We use the sentence embedding method SimCSE in our experiments. We use the \textit{sup-simcse-bert-base-uncased} model as the base encoder.

\noindent\textbf{Graph Encoder (\AMRSIMNAME)}: During training, we set the positive ratio to 4/16, meaning each batch of 16 contains 4 positive graph pairs and 12 negative pairs. Specifically, we sampled 4 graphs and generated one positive pair and three negative pairs for each graph. The transformer parameters were initialized using the uncased BERT base model \cite{devlin-etal-2019-bert}, while the graph adapter parameters were initialized randomly. Hyperparameters were set as follows: 1 epoch, learning rate as 1e-5, dropout rate as 0.1, and graph adapter size as 128. We experimented with sequence length of 128 for SemEval and 256 for the other three datasets. The training was done using NVIDIA Quadro RTX 8000.

\noindent\textbf{Supervised RE Model (PURE)}: To maintain consistency across datasets, we use a single-sentence setup for Semeval, as it is a sentence-level relation extraction dataset. For pre-trained language models (PLMs), we follow PURE by using scibert-scivocab-uncased \cite{beltagy-etal-2019-scibert} as the base encoder for SciERC and bert-base-uncased \cite{devlin-etal-2019-bert} for the other three datasets. We also adhere to the hyperparameters specified in their paper.

\section{Results of All \MNAME~configurations}
Table \ref{table:allresults} shows the results for all the configurations in our experiments.


\end{document}